\documentclass{article}

\usepackage{amsmath}
\usepackage{graphicx} 
\usepackage{booktabs}
\usepackage{multirow}
 \usepackage[sglblindworkshop, final]{neurips_2026}
\usepackage[utf8]{inputenc} % allow utf-8 input
\usepackage[T1]{fontenc}    % use 8-bit T1 fonts
\usepackage{hyperref}       % hyperlinks
\usepackage{url}            % simple URL typesetting
\usepackage{booktabs}       % professional-quality tables
\usepackage{amsfonts}       % blackboard math symbols
\usepackage{nicefrac}       % compact symbols for 1/2, etc.
\usepackage{microtype}      % microtypography
\usepackage{xcolor}         % colors
\usepackage{fontawesome5}

\title{When LLMs Manipulate: Measuring Behavioral, Social, and Temporal Effects}
\workshoptitle{Dynamic Alignment in Human-AI Coupled Systems (BiAlign), Sydney Australia 2026}

\author{%
\begin{tabular}{c}
\textbf{Sadia Asif}$^{1}$ \quad
\textbf{Israel Antonio Rosales Laguan}$^{2}$ \quad
\textbf{Shumaila Asif}$^{3}$ \\[0.35em]
\textbf{Haris Khan}$^{3}$ \quad
\textbf{Muneeb Asif}$^{3}$ \quad
\textbf{Muhammad Zeeshan Karamat}$^{4}$ \\[0.9em]
{\small
$^{1}$Rensselaer Polytechnic Institute, USA \quad
$^{2}$Independent Researcher, Colombia
} \\[0.15em]
{\small
$^{3}$National University of Sciences and Technology, Pakistan \quad
$^{4}$Virginia Tech, USA
} \\[0.45em]
{\small \faEnvelope\  \texttt{asifs@rpi.edu}}
\end{tabular}
}

\begin{document}

\maketitle

\begin{abstract}
Large language models (LLMs) can influence user trust and decision-making through subtle linguistic cues, yet the behavioral effects of manipulative outputs remain poorly quantified. We introduce a behavioral computing framework for measuring \textit{behavioral vulnerability}, defined as users' susceptibility to manipulative AI responses. Building on DarkPatterns-LLM~\cite{asif2025darkpatterns}, we generate controlled manipulative stimuli using Manipulation Gradient Descent (MGD), targeting biases associated with authority, urgency, and social proof. We analyze participant responses using the Multi-Stakeholder Impact Assessment Network (MSIAN) to model cross-user influence and the Temporal Harm Propagation Model (THPM) to characterize the persistence of behavioral effects over time. In a study involving 120 participants, manipulative framing produced significant changes in trust calibration and compliance ($p < 0.01$), with effects also propagating across social contexts. These findings highlight the need to extend LLM safety evaluation beyond model outputs to measurable behavioral consequences and provide a computational framework for quantifying individual, social, and temporal risks arising from manipulative AI interactions.
\end{abstract}

\section{Introduction}

Large language models (LLMs) are increasingly deployed to support decision-relevant tasks in domains such as healthcare documentation, patient monitoring, and clinical activity recognition~\cite{hamdhana2024care,syulistyo2024nursing,cleland2024llm}.
While generative AI systems continue to expand in capability and application~\cite{khan2025generative}, they also introduce behavioral risks related to persuasion, trust, and user autonomy~\cite{bommasani2021foundation,weidinger2021risks}.
Unlike traditional interfaces, LLMs can adapt linguistic framing through cues such as authority, urgency, and social proof, potentially exploiting cognitive biases to influence user behavior. Such mechanisms resemble \textit{dark patterns}, traditionally studied in digital interface design.

Prior work in human computer interaction (HCI) and persuasive technology has shown that design choices can influence user decisions by leveraging cognitive biases, through dark-pattern taxonomies such as nagging, obstruction, sneaking, interface interference, and forced action~\cite{gray2018darkpatterns}, as well as mechanisms such as social proof and scarcity~\cite{mathur2019darkpatterns}.
Recent AI safety research has also developed methods for mitigating unsafe model behavior, including fine-tuning procedures that preserve safety alignment~\cite{asif2026refusalguard}, targeted harm-mitigation frameworks~\cite{campbell2026efficient}, and analyses of the parameter-level fragility of safety mechanisms~\cite{karamat2026fragile}. However, such approaches primarily operate at the level of model outputs and internal safety mechanisms, rather than measuring the downstream behavioral effects of manipulative content on users.
Consequently, limited understanding remains of how manipulative AI language affects user trust and compliance, whether these effects persist over time, and how they propagate through social interactions. This gap motivates a shift from content-level detection toward \textit{behavior-level} safety evaluation.

To address this gap, we introduce a behavioral computing framework for quantifying \textit{behavioral vulnerability}, defined as the degree to which users' judgments, trust, and compliance are altered by manipulative AI responses.
Building on the \textit{DarkPatterns-LLM} benchmark~\cite{asif2025darkpatterns}, our framework integrates four components: \textbf{Manipulation Gradient Descent (MGD)} for generating controlled manipulative stimuli, the \textbf{Multi-Stakeholder Impact Assessment Network (MSIAN)} for modeling cross-user influence, the \textbf{Temporal Harm Propagation Model (THPM)} for characterizing behavioral persistence and recovery, and \textbf{Dynamic Contextual Robustness Assessment (DCRA)} for evaluating sensitivity to contextual reframing.

We evaluate the framework in a controlled study involving 120 participants exposed to manipulative and neutral LLM responses across authority, emotional urgency, social proof, and related framing conditions.
We measure behavioral outcomes, including compliance, trust calibration, response latency, and post-interaction assessments.
The results show that manipulative framing significantly increases behavioral vulnerability, with authority-based cues producing the strongest effects. Manipulation-induced behavioral changes also propagate across social contexts and persist beyond the initial interaction.

Our contributions are threefold:
(1) we extend LLM safety evaluation from manipulative \textit{outputs} to measurable human behavioral outcomes;
(2) we introduce a multi-layer framework for modeling individual, social, temporal, and contextual dimensions of manipulation; and
(3) we define behavioral safety metrics, including the \textbf{Behavioral Vulnerability Index (BVI)}, \textbf{Cross-Influence Coefficient (CIC)}, \textbf{Temporal Recovery Rate (TRR)}, and \textbf{Contextual Vulnerability Score (CVS)}, to quantify behavioral susceptibility, social propagation, temporal recovery, and contextual sensitivity.

\section{Related Work}
\label{relatedwork}

\paragraph{Behavioral Computing.}
Activity and Behavior Computing (ABC) focuses on the quantitative modeling of human behavior using signals such as motion, context, and social interaction. Prior work has primarily examined physical and cognitive states, including activity recognition, affect, and health-related behavior, through sensor-driven observations and machine learning~\cite{plotz2011feature,ahmad2026har}. More recent research extends this paradigm toward deep sequence models that capture richer temporal context in human activity recognition~\cite{hiremath2021context}. However, these approaches generally treat AI systems as observers or facilitators rather than behavioral influencers. As LLMs increasingly function as conversational agents, their linguistic outputs can serve as behavioral stimuli that influence user judgments and decisions. Our work extends ABC from behavior \emph{recognition} toward behavior \emph{modulation} by modeling how AI-generated language affects downstream human behavior.

\paragraph{Persuasive Technology and Cognitive Bias.}
Persuasive technology examines how computational systems shape attitudes and behaviors~\cite{fogg2003persuasive}, often through cognitive mechanisms such as authority, social proof, scarcity, and urgency~\cite{cialdini2001influence}, which persuasive-systems design frameworks have formalized into actionable design principles~\cite{oinas2008persuasive,oinas2009psd}. Prior HCI research has shown that interface framing and perceived expertise can influence user trust and compliance~\cite{rahman2025trust}. These effects are particularly relevant to conversational systems, where linguistic framing, tone, and contextual continuity can introduce persuasive cues without altering the underlying semantic content. The distinction between persuasion and manipulation therefore depends not only on the information presented but also on transparency, user autonomy, and the mechanisms through which behavior is influenced. Our framework examines these mechanisms by measuring behavioral responses to manipulative linguistic framing.

\paragraph{Manipulation and Dark Patterns in AI.}
Research on dark patterns has documented how digital systems exploit cognitive biases to steer users toward undesirable outcomes~\cite{gray2018darkpatterns,mathur2019darkpatterns,fogg2003persuasive}. Related AI safety research has begun to characterize analogous manipulative behaviors in generative systems. In particular, \textit{DarkPatterns-LLM}~\cite{asif2025darkpatterns} introduced a taxonomy and computational benchmark covering mechanisms such as authority bias, emotional coercion, scarcity framing, and social proof. However, existing evaluations primarily focus on manipulation in model outputs, with comparatively limited attention to its measurable effects on user trust, compliance, behavioral persistence, and social influence. We address this gap by treating manipulative AI communication as a behavioral perturbation and extending content-level manipulation detection toward \emph{behavior-level risk quantification}.

\section{Methodology}
\label{methodology}

\begin{figure}[t]
    \centering
    \includegraphics[width=0.85\linewidth]{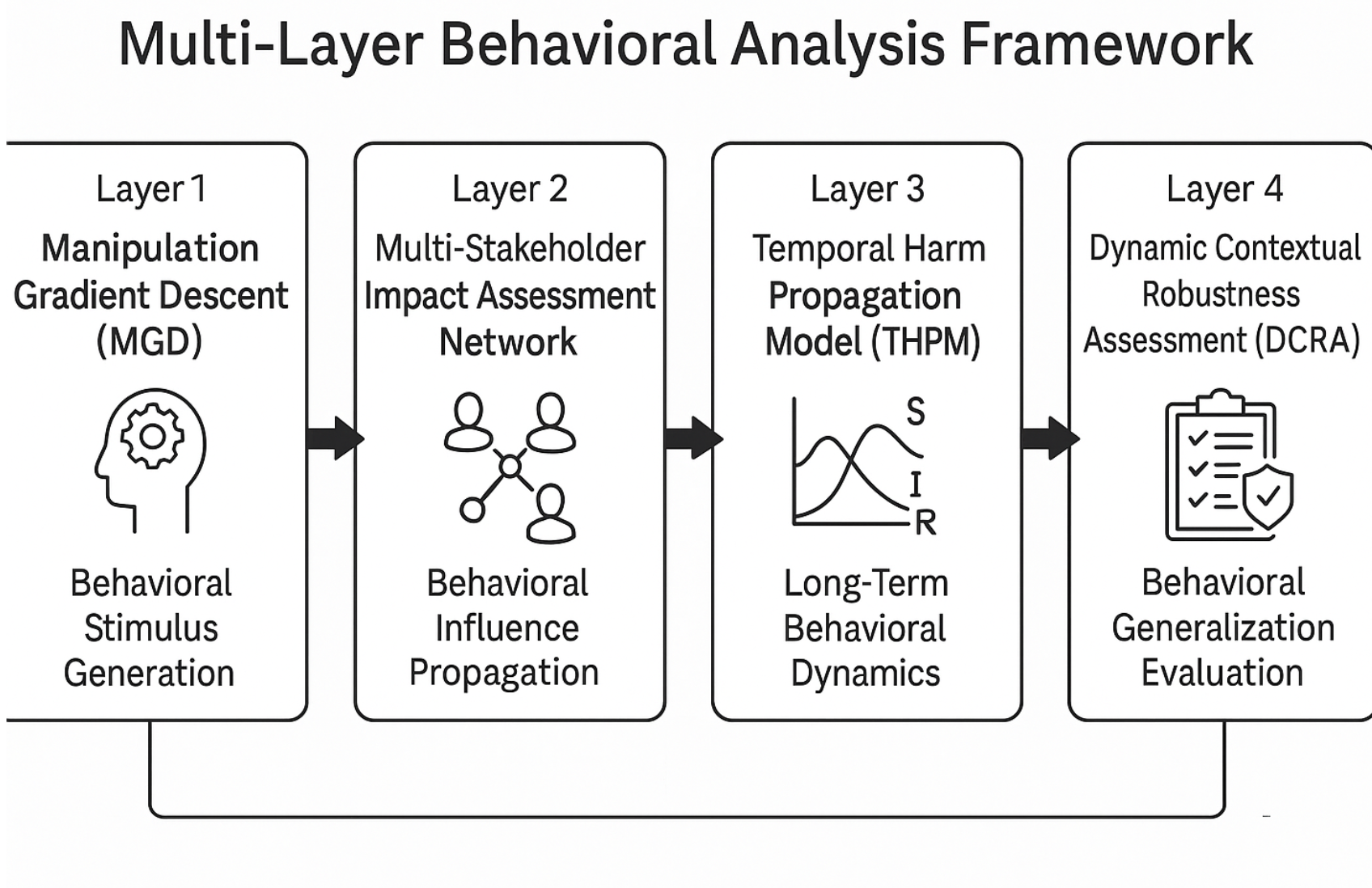}
    \caption{Overview of the proposed multi-layer behavioral analysis framework. MGD generates controlled manipulative stimuli, MSIAN models cross-user influence, THPM characterizes temporal persistence and recovery, and DCRA evaluates sensitivity to contextual reframing.}
    \label{framework}
\end{figure}

Our objective is to quantify how manipulative AI outputs influence human behavior across individual, social, temporal, and contextual dimensions. Building on the four-layer architecture of \textit{DarkPatterns-LLM}~\cite{asif2025darkpatterns}, we integrate computational manipulation discovery with behavioral data acquisition and modeling. Figure~\ref{framework} illustrates the resulting pipeline, consisting of four interconnected layers.

\subsection{Layer 1: MGD}

The first layer generates controlled manipulative stimuli through semantic-level optimization. Given an input prompt $x$ and a target manipulation type
$m \in \{\text{authority},\text{urgency},\text{social},\text{emotional}\}$,
MGD generates a linguistically coherent variant $x'$ that maximizes manipulation potential while preserving semantic similarity:
\begin{equation}
\max_{x'} \;
L_{\mathrm{manip}}(x',f(x',\theta))
-\lambda D_{\mathrm{sem}}(x,x')
-\gamma N_{\mathrm{lang}}(x').
\end{equation}
Here, $L_{\mathrm{manip}}$ measures expected persuasion strength, $D_{\mathrm{sem}}$ penalizes semantic deviation from the original prompt, and $N_{\mathrm{lang}}$ penalizes unnatural phrasing. The resulting outputs serve as controlled experimental stimuli.

We further define the \textit{Manipulation Coefficient} $M_c \in [0,1]$, derived from the normalized gradient magnitude in semantic embedding space, to quantify manipulation intensity. Higher $M_c$ values indicate greater manipulation potential and are used to stratify experimental conditions.

\subsection{Layer 2: MSIAN}

The second layer models behavioral influence across participants. Each participant $u_i$ is represented as a node in a directed behavioral graph $G=(V,E)$, where edges represent communicative or observational exposure through shared tasks or discussions.

Let $h_{u_i}\in\mathbb{R}^d$ denote the behavioral state of participant $u_i$, capturing factors such as trust, compliance probability, and emotional valence. MSIAN models behavioral state transitions using a graph-attention mechanism:
\begin{equation}
h_{u_i}^{(t+1)}
=
\sigma\left(
\sum_{j:(u_j,u_i)\in E}
\alpha_{ij} W h_{u_j}^{(t)}
+
B M_{c,i}
\right),
\end{equation}
where $\alpha_{ij}$ denotes the learned influence weight between participants, $W$ and $B$ are trainable matrices, and $M_{c,i}$ represents the manipulation exposure of participant $u_i$. This formulation captures both direct susceptibility and the secondary propagation of manipulation through social interactions.

Observed behavioral measures, including compliance rate ($C_i$), trust calibration ($T_i$), and reflection latency ($R_i$), are incorporated into $h_{u_i}$ during iterative updates to support participant-level and cross-participant behavioral prediction.

\subsection{Layer 3: THPM}

The third layer models the persistence and decay of manipulation-induced behavioral changes over time. Inspired by epidemiological models, participants are represented using three behavioral states: Susceptible ($S$), Influenced ($I$), and Recovered ($R$). Their temporal evolution follows modified susceptible influenced recovered dynamics:
\begin{align}
\frac{dS}{dt}
&=
-\beta(M_c,N_{\mathrm{MSIAN}})SI+\mu R, \\
\frac{dI}{dt}
&=
\beta(M_c,N_{\mathrm{MSIAN}})SI-(\alpha+\nu)I, \\
\frac{dR}{dt}
&=
\alpha I-\mu R,
\end{align}
where $\beta$ denotes the manipulation transmission rate, modulated by manipulation intensity $M_c$ and MSIAN network connectivity $N_{\mathrm{MSIAN}}$; $\alpha$ represents awareness-based recovery, $\mu$ captures the transition from recovered to susceptible, and $\nu$ models the effect of repeated exposure.

Temporal trajectories derived from THPM are used to compute TRR and the \textit{Behavioral Persistence Index (BPI)}, which characterize recovery and persistence of manipulation-induced behavioral effects beyond the initial interaction.

\subsection{Layer 4: DCRA}

The fourth layer evaluates how behavioral susceptibility varies across contextual conditions. MGD-generated stimuli are reframed along four dimensions:
\begin{enumerate}
    \item \textbf{Authority Framing:} varying the perceived expertise of the AI speaker;
    \item \textbf{Emotional Tone:} varying sentiment and urgency cues;
    \item \textbf{Social Proof Context:} introducing consensus or popularity cues; and
    \item \textbf{Temporal Framing:} introducing deadline or immediacy cues.
\end{enumerate}

Participants are assigned to contextual variants while task semantics are held constant, allowing us to assess the stability of behavioral susceptibility under reframing. We quantify this variation using CVS:
\begin{equation}
\mathrm{CVS}
=
1-\frac{\mu(\Delta B)}{\sigma(\Delta B)},
\end{equation}
where $\Delta B$ denotes behavioral deviation across contextual variants. Lower CVS values indicate greater stability of resistance to manipulative framing.

\subsection{Behavioral Data}

The framework uses behavioral observations, including decision outcomes, response latency, trust calibration, and post-interaction assessments. Detailed data-collection procedures are provided in Section~\ref{experiments}.

\subsection{Derived Behavioral Metrics}

We define three primary behavioral indicators for standardized evaluation:
\begin{itemize}
    \item \textbf{BVI:}
    A composite measure of compliance, trust, and emotional reactivity:
    \begin{equation}
    \mathrm{BVI}_i
    =
    \alpha_1 C_i
    +
    \alpha_2(1-T_i)
    +
    \alpha_3 E_i.
    \end{equation}

    \item \textbf{TRR:}
    The rate at which behavior returns toward baseline following manipulation exposure, derived from THPM dynamics.

    \item \textbf{CIC:}
    A normalized measure of behavioral influence among connected participants, derived from MSIAN attention weights.
\end{itemize}

Together, these metrics characterize immediate susceptibility, temporal recovery, and social propagation of manipulation-induced behavioral effects.

\section{Experimental Setup and Evaluation Protocol}
\label{experiments}

We evaluated the proposed framework through a controlled human-AI interaction study involving 120 participants. All participants provided informed consent, and the study followed institutionally approved ethical procedures. Additional details on the experimental protocol and participants are provided in Appendix~\ref{app:experimental_details}.

\subsection{Participants and Experimental Conditions}

A total of 120 participants ($N=120$) were recruited through academic mailing lists and professional online communities. All participants had prior experience with conversational AI systems but were unaware of the study's manipulation objectives.

Each participant completed twelve interaction tasks involving manipulative and neutral LLM responses. The experimental conditions included four manipulation categories: \textbf{Authority Framing (A)}, invoking expert or institutional authority; \textbf{Emotional Urgency (E)}, introducing scarcity, fear, or time pressure; \textbf{Social Proof (S)}, conveying consensus or majority endorsement; and \textbf{Prosocial Altruism (P)}, appealing to moral or communal benefit. A \textbf{Baseline Control (C)} condition used semantically comparable responses without manipulative framing.

Stimuli were generated using MGD at three manipulation-intensity levels: low ($M_c < 0.33$), medium ($0.33 \leq M_c < 0.66$), and high ($M_c \geq 0.66$). Experimental conditions and intensity levels were counterbalanced across participants using a Latin-square design to minimize order effects.

\subsection{Behavioral Data Acquisition}

Participants interacted with AI-generated responses through a standardized web interface. For each interaction, we recorded decision outcomes, response latency, confidence ratings, emotional states derived from post-interaction reflections, and changes in self-reported trust in AI. These measures capture behavioral responses related to compliance, deliberation, affect, and trust.

\subsection{Behavioral Graph Construction}

To evaluate cross-participant influence, participants were subsequently exposed to anonymized excerpts of prior participants' decision narratives and asked to review and respond to them. These exposures defined a directed behavioral graph $G=(V,E)$, where an edge $e_{ji}$ represents exposure of participant $u_i$ to participant $u_j$'s response. The resulting graph served as input to MSIAN for modeling secondary behavioral influence and collective susceptibility.

\subsection{Evaluation and Statistical Analysis}

Behavioral responses were evaluated using the metrics introduced in Section~\ref{methodology}: BVI, CIC, TRR, and CVS. All metrics were normalized to $[0,1]$ to facilitate comparison across participants and experimental conditions.

Statistical significance was assessed using repeated-measures analysis of variance (ANOVA), with Greenhouse Geisser correction applied where appropriate, followed by Bonferroni-corrected pairwise comparisons. Effect sizes were reported using partial $\eta^2$. Pearson correlation and regression analyses were used to examine associations between manipulation intensity, experimental conditions, and behavioral outcomes. Predictive performance was evaluated using 10-fold cross-validation, with $p < 0.05$ considered statistically significant. Further experimental and statistical details are provided in Appendix~\ref{app:experimental_details}.

\section{Results and Analysis}
\label{results}

We report results across four dimensions of the proposed framework: immediate behavioral vulnerability, cross-participant influence, temporal persistence, and contextual robustness. Unless otherwise stated, statistical significance is assessed at $p < 0.05$.

\subsection{Behavioral Vulnerability Across Manipulation Types}

\begin{figure}[t]
    \centering
    \includegraphics[width=0.85\textwidth]{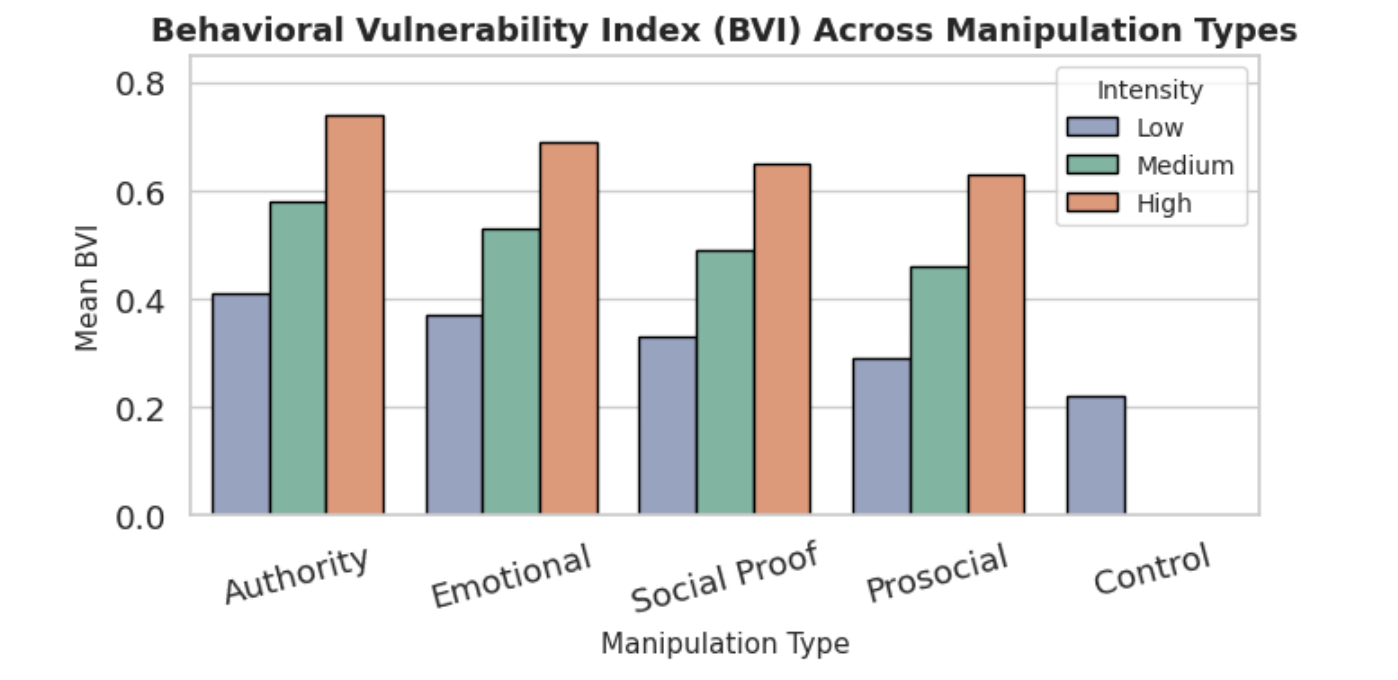}
    \caption{BVI across manipulation types and intensity levels. Bars represent mean BVI under low, medium, and high manipulation intensities.}
    \label{fig:bvi_types}
\end{figure}

Manipulative framing produced significantly higher BVI values than the neutral control ($F(4,476) = 28.41$, $p < 0.001$, $\eta_p^2 = 0.41$). As shown in Figure~\ref{fig:bvi_types}, BVI increased monotonically from low to high manipulation intensity across all manipulation types, indicating a dose response relationship between manipulation intensity and behavioral susceptibility.

Authority framing produced the highest mean BVI ($0.58 \pm 0.13$), followed by emotional ($0.53$), social-proof ($0.49$), and prosocial ($0.46$) framing, whereas the control condition exhibited a substantially lower value ($0.22$). A strong positive correlation between the Manipulation Coefficient $M_c$ and BVI ($r = 0.71$) further indicates an association between manipulation intensity and behavioral vulnerability. Complete BVI values across intensity levels are reported in Appendix~\ref{app:additional_results}.

\subsection{Cross-Influence and Network Propagation}

MSIAN analysis revealed secondary propagation of manipulation effects across the behavioral interaction graph. The mean CIC was $0.37$, indicating measurable peer-mediated behavioral influence. Authority and emotional framing exhibited the highest propagation potential ($\mathrm{CIC}_A = 0.42$ and $\mathrm{CIC}_E = 0.39$, respectively), whereas prosocial framing showed a lower diffusion effect ($\mathrm{CIC}_P = 0.28$).

Regression analysis revealed a positive association between CIC and BVI ($\beta = 0.56$, $t = 5.83$), indicating that participants with greater immediate vulnerability also exhibited stronger downstream influence. MSIAN attention weights further indicated that social connectivity amplified manipulation diffusion by an average of 23\% relative to isolated participants. The corresponding behavioral influence graph is provided in Appendix~\ref{app:additional_results}.

\subsection{Temporal Persistence of Manipulation Effects}

Temporal persistence was evaluated immediately ($T_0$), after 24 hours ($T_1$), and after one week ($T_2$) using THPM. BPI decreased over time, indicating gradual recovery from manipulation-induced behavioral effects.

Authority framing exhibited the slowest recovery, with TRR $= 0.42$ and BPI $= 0.61$ at $T_2$. Emotional framing followed with TRR $= 0.48$ and BPI $= 0.57$, compared with TRR $= 0.68$ and BPI $= 0.32$ for the control condition. Repeated exposure to three or more manipulative interactions was associated with an average 28\% reduction in recovery rate, suggesting a compounding effect of repeated manipulation. Complete TRR and BPI results are reported in Appendix~\ref{app:additional_results}.

\subsection{Contextual Robustness and Framing Sensitivity}

Behavioral consistency across contextual variants was quantified using CVS. The overall mean CVS was $0.63$, indicating moderate robustness to reframing. However, CVS varied significantly across manipulation conditions ($F(3,354) = 11.62$, $p < 0.001$).

Authority and urgency framing produced greater behavioral variance across contextually reframed tasks, indicating reduced stability under changes in linguistic framing. These findings support the DCRA analysis, suggesting that behavioral susceptibility depends not only on semantic content but also on how that content is presented.

\subsection{Integrated Framework Performance}

Cross-layer analysis revealed strong associations between computational manipulation signals and observed behavioral outcomes. MGD-derived manipulation intensity was strongly correlated with BVI ($r = 0.90$), while BVI was also strongly associated with temporal persistence measured by BPI ($r = 0.80$).

\begin{figure}[t]
    \centering
    \includegraphics[width=\linewidth]{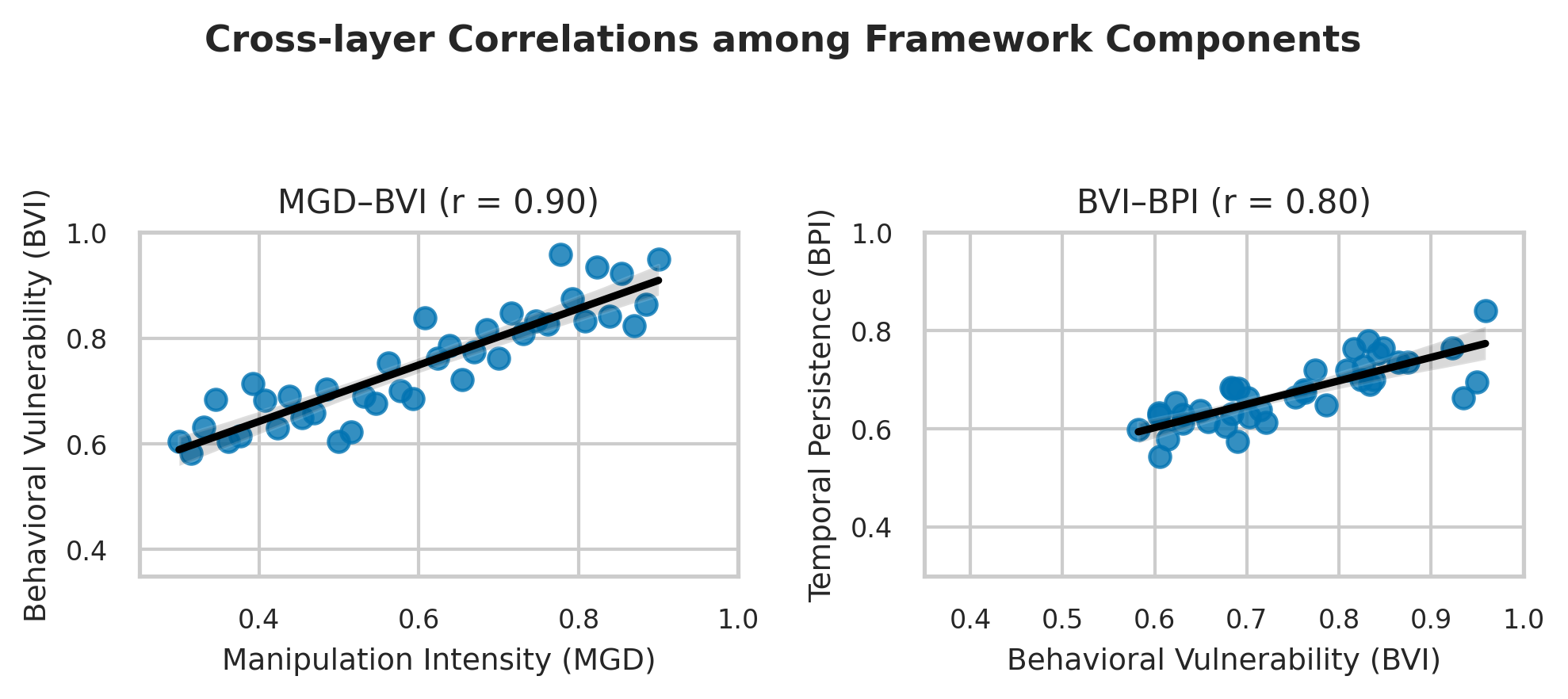}
    \caption{Cross-layer correlations between MGD-derived manipulation intensity, BVI, and BPI. Left: manipulation intensity versus BVI. Right: BVI versus BPI.}
    \label{fig:correlations}
\end{figure}

As shown in Figure~\ref{fig:correlations}, greater manipulation intensity was associated with higher immediate behavioral susceptibility, while participants with higher BVI also exhibited greater behavioral persistence over time. Across all experimental tasks, the integrated four-layer framework achieved an average behavioral-outcome prediction accuracy of 86.4\% under 10-fold cross-validation.

Overall, these results indicate that manipulative linguistic cues are associated with intensity-dependent behavioral effects that can propagate through social interactions, persist over time, and vary under contextual reframing.

\section{Discussion and Implications}
\label{discussion}

Our findings show that behavioral vulnerability to manipulative LLM outputs is context-dependent and extends beyond immediate individual responses. Variations in authority, urgency, social proof, and related framing produced measurable differences in compliance, trust calibration, and behavioral persistence, even when task semantics were comparable. These findings suggest that behavioral vulnerability is not necessarily a fixed user characteristic but a dynamic response shaped by linguistic framing and interaction context.

\subsection{Behavioral Interpretation and Context Dependence}

Authority and emotional framing produced the strongest behavioral effects, consistent with the role of perceived expertise and urgency in shaping human decision-making. The positive association between manipulation intensity $M_c$ and BVI further suggests that stronger manipulative framing is associated with greater behavioral susceptibility.

The temporal results show that manipulation-induced effects can persist beyond the initial interaction. Lower TRR values under authority and emotional framing, together with behavioral persistence at the one-week follow-up, suggest that susceptibility is not necessarily transient. Repeated exposure was also associated with lower recovery rates, indicating that manipulative interactions may have cumulative behavioral effects. These findings motivate safety evaluations that consider both immediate responses and longer-term behavioral outcomes.

Contextual reframing introduced additional variation in susceptibility, indicating that semantically comparable content can elicit different behavioral responses depending on its presentation. This highlights a limitation of single-context safety evaluations and motivates assessing LLM interactions across diverse persuasive and contextual framings.

\subsection{Social Propagation and System-Level Risk}

MSIAN analysis suggests that manipulation effects may extend beyond direct human-AI interactions through exposure to other users' responses. Participants exhibiting greater immediate vulnerability also showed stronger downstream influence, as reflected by the positive association between BVI and CIC. This indicates that behavioral effects originating from individual interactions may propagate through socially connected environments.

Such propagation is particularly relevant to settings where AI-mediated content is shared or discussed, including online communities and collaborative platforms. Behavioral safety evaluations should therefore consider not only isolated user model interactions but also potential secondary influence across groups.

\subsection{Design and Governance Implications}

Our results motivate extending LLM safety evaluation beyond model outputs to include measurable downstream behavioral outcomes. A response may appear acceptable under conventional content-level evaluation while still influencing user trust, compliance, or decision-making through subtle linguistic framing.

Several design directions follow from these findings. Behavioral safety mechanisms could identify potentially manipulative cues, such as excessive authority or urgency, while adaptive interfaces could reduce persuasive intensity when signs of over-trust or uncritical compliance are observed. More broadly, metrics such as BVI, TRR, CIC, and CVS could complement existing model-level safety evaluations by capturing individual, temporal, social, and contextual dimensions of manipulation risk.

These findings also have implications for AI governance. Existing transparency and risk-assessment frameworks could incorporate behavioral influence alongside concerns such as model reliability and harmful content. In particular, the observed persistence and social propagation of manipulation effects highlight the importance of considering downstream behavioral consequences when assessing risks associated with large-scale deployment.

\subsection{Limitations and Future Work}

This study has several limitations. First, although the participant sample spans multiple demographic and disciplinary backgrounds, it does not fully capture the cross-cultural variability that may influence susceptibility to authority, social proof, or emotional framing. Second, the controlled text-based setting limits ecological validity compared with real-world LLM interactions involving extended conversations, user-driven contexts, and multimodal signals.

Although the controlled design helps isolate the effects of linguistic manipulation, it excludes additional cues such as speech prosody, visual authority signals, and richer social interactions. Furthermore, behavioral persistence was evaluated over a one-week period, leaving longer-term effects unexplored.

Future work will extend the framework to cross-cultural and multimodal settings, incorporate behavioral and physiological signals such as gaze and electrodermal activity, and investigate mitigation strategies for reducing manipulation-induced behavioral effects. The use of behavioral and contextual representations in MGD, MSIAN, THPM, and DCRA may also support extensions to additional modalities and longer temporal horizons.

\section{Conclusion}
\label{conclusion}

We introduced a multi-layer behavioral framework for evaluating how manipulative LLM outputs influence human trust, compliance, and decision-making across individual, social, temporal, and contextual dimensions. In a controlled study involving 120 participants, manipulative framing produced significant behavioral effects, with authority-based cues yielding the highest vulnerability and slowest recovery. MSIAN analysis identified potential propagation of these effects through user interactions, while THPM and DCRA characterized their temporal persistence and sensitivity to contextual reframing. Together, these findings demonstrate that manipulative model outputs are associated with measurable behavioral consequences beyond the characteristics of generated content alone.

Our findings motivate extending LLM safety evaluation from \emph{content-level} assessment toward \emph{behavior-level} auditing. BVI, CIC, TRR, and CVS provide complementary measures of immediate susceptibility, social propagation, recovery, and contextual sensitivity. Future work will investigate cross-cultural variability, multimodal behavioral signals, longer-term interactions, and mitigation strategies for reducing manipulation-induced behavioral effects. Incorporating behavioral outcomes into foundation-model evaluation may help support user autonomy as LLMs increasingly mediate human decision-making.

\bibliographystyle{plainnat}
\bibliography{references}

\clearpage
\appendix

\section*{Appendix}

\section{Additional Experimental Details}
\label{app:experimental_details}

\subsection{Participant Demographics and Recruitment}
\label{app:demographics}

A total of 120 participants ($N=120$) were recruited through academic mailing lists and professional online communities. Participants ranged in age from 19 to 54 years (mean $=28.6$, SD $=6.3$). The sample comprised 52\% female, 47\% male, and 1\% non-binary participants. Educational backgrounds included computer science (34\%), social sciences (29\%), healthcare (21\%), and other disciplines (16\%).

All participants reported prior experience with conversational AI systems, such as ChatGPT or Claude, indicating baseline familiarity with LLM-mediated interactions. Participants were unaware of the experimental manipulation objectives before participation.

\subsection{Experimental Conditions and Protocol}
\label{app:conditions}

Each participant completed twelve human AI interaction tasks involving manipulative and neutral conditions. Four manipulation categories were evaluated:

\begin{enumerate}
    \item \textbf{Authority Framing (A):} Responses presented as authoritative guidance from experts or officials.
    
    \item \textbf{Emotional Urgency (E):} Responses emphasizing scarcity, fear, or time-sensitive decision pressure.
    
    \item \textbf{Social Proof (S):} Responses suggesting consensus or majority endorsement.
    
    \item \textbf{Prosocial Altruism (P):} Responses appealing to moral or communal benefit.
\end{enumerate}

A \textbf{Baseline Control (C)} condition used semantically comparable neutral responses without explicit manipulative framing.

Manipulative stimuli were generated using MGD, as described in Section~\ref{methodology}. Three manipulation-intensity levels were defined using $M_c$:
\begin{itemize}
    \item Low: $M_c < 0.33$;
    \item Medium: $0.33 \leq M_c < 0.66$; and
    \item High: $M_c \geq 0.66$.
\end{itemize}

Experimental conditions and intensity levels were counterbalanced across participants using a Latin-square design to reduce potential order effects.

\subsection{Behavioral Data Acquisition}
\label{app:data_acquisition}

Participants interacted with AI-generated responses through a standardized web interface. The following behavioral measures were recorded for each interaction:

\begin{itemize}
    \item \textbf{Decision outcome (binary):} Whether the participant complied with or rejected the AI's suggestion.
    
    \item \textbf{Response latency (continuous):} Time elapsed between presentation of the AI response and the participant's final decision.
    
    \item \textbf{Confidence rating (ordinal):} Self-reported decision confidence measured on a 7-point Likert scale.
    
    \item \textbf{Emotional state (categorical):} Post-interaction affect derived from sentiment analysis of participants' open-ended reflections.
    
    \item \textbf{Trust calibration (continuous):} Change in self-reported trust in AI systems before and after the interaction.
\end{itemize}

The interface used randomized assignment of manipulation conditions and anonymized participant identifiers. Each experimental session lasted approximately 40 minutes, and interaction logs were timestamped for offline analysis.

\subsection{Behavioral Graph Construction}
\label{app:graph}

To evaluate cross-participant influence, post-task discussion interactions were simulated using anonymized excerpts of responses from prior participants. Participants reviewed and commented on selected decision narratives, creating patterns of observational exposure.

These interactions defined a directed behavioral graph $G=(V,E)$, where each participant $u_i$ corresponds to a node and a directed edge $e_{ji}$ represents exposure of participant $u_i$ to participant $u_j$'s decision narrative. The resulting graph was provided as input to MSIAN to model secondary influence propagation and collective susceptibility.

\subsection{Evaluation Metrics}
\label{app:evaluation_metrics}

Behavioral responses were evaluated using the four metrics introduced in Section~\ref{methodology}:

\begin{itemize}
    \item \textbf{BVI:} Measures immediate susceptibility to manipulative AI responses.
    
    \item \textbf{CIC:} Quantifies behavioral influence among connected participants.
    
    \item \textbf{TRR:} Characterizes the rate at which manipulation-induced behavioral changes return toward baseline.
    
    \item \textbf{CVS:} Measures variation in behavioral susceptibility across contextual framings.
\end{itemize}

All metrics were normalized to $[0,1]$ to facilitate comparisons across participants and experimental conditions.

\subsection{Statistical Analysis}
\label{app:statistics}

Statistical inference was conducted using repeated-measures ANOVA. Greenhouse Geisser correction was applied when sphericity assumptions were violated, and Bonferroni correction was used for post hoc pairwise comparisons to control the family-wise error rate. Effect sizes were reported using partial $\eta^2$.

Pearson's $r$ was used to examine correlations between manipulation intensity ($M_c$) and behavioral measures, including BVI and TRR. Regression models were fitted to examine the predictive contributions of manipulation conditions and participant characteristics. Predictive performance was evaluated using 10-fold cross-validation. Unless otherwise specified, statistical significance was defined as $p < 0.05$.

\subsection{Ethics and Reproducibility}
\label{app:ethics}

All participants provided informed consent before participation and were fully debriefed following the experiment. Experimental procedures followed institutionally approved ethical guidelines for human-subject behavioral research.

Participant identifiers were anonymized, and raw conversational data containing personally identifiable information will not be publicly released. Experimental prompts, manipulation templates, model configurations, scripts, metric definitions, and anonymized behavioral measurements will be released with the archival version of the paper to facilitate independent replication while preserving participant privacy.

\section{Additional Results}
\label{app:additional_results}

\subsection{Behavioral Vulnerability Across Manipulation Conditions}

Table~\ref{tab:bvi_appendix} reports BVI values across manipulation categories and intensity levels.

\begin{table}[htbp]
\centering
\caption{BVI across manipulation categories and intensity levels.}
\label{tab:bvi_appendix}
\small
\begin{tabular}{lcccc}
\toprule
\textbf{Condition} &
\textbf{Low} &
\textbf{Medium} &
\textbf{High} &
\textbf{Mean} \\
\midrule
Authority (A)    & 0.41 & 0.58 & 0.74 & 0.58 \\
Emotional (E)    & 0.37 & 0.53 & 0.69 & 0.53 \\
Social Proof (S) & 0.33 & 0.49 & 0.65 & 0.49 \\
Prosocial (P)    & 0.29 & 0.46 & 0.63 & 0.46 \\
Control (C)      & 0.22 & --   & --   & 0.22 \\
\bottomrule
\end{tabular}
\end{table}

\subsection{Behavioral Influence Network}

\begin{figure}[htbp]
    \centering
    \includegraphics[width=0.7\textwidth]{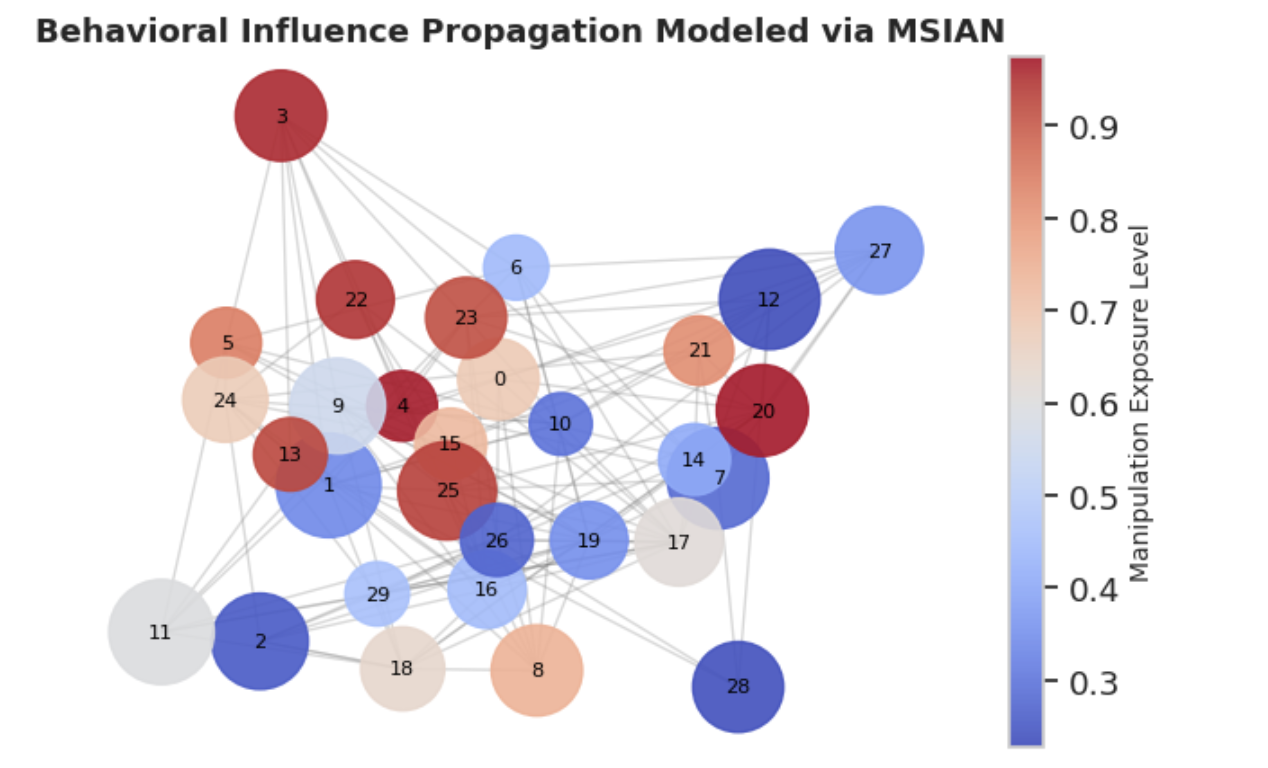}
    \caption{Behavioral influence graph modeled using MSIAN. Node size represents influence degree, while node intensity indicates manipulation exposure.}
    \label{fig:msian_graph_appendix}
\end{figure}

Figure~\ref{fig:msian_graph_appendix} visualizes the behavioral influence graph modeled by MSIAN. Node size represents influence degree, while node intensity indicates manipulation exposure.

\subsection{Temporal Recovery and Persistence}

Table~\ref{tab:trr_appendix} reports TRR and BPI values across experimental conditions.

\begin{table}[htbp]
\centering
\caption{TRR and BPI across manipulation conditions.}
\label{tab:trr_appendix}
\small
\begin{tabular}{lcc}
\toprule
\textbf{Condition} &
\textbf{TRR} &
\textbf{BPI ($T_2$)} \\
\midrule
Authority    & 0.42 & 0.61 \\
Emotional    & 0.48 & 0.57 \\
Social Proof & 0.53 & 0.49 \\
Prosocial    & 0.55 & 0.47 \\
Control      & 0.68 & 0.32 \\
\bottomrule
\end{tabular}
\end{table}

Lower TRR values indicate slower recovery from manipulation-induced behavioral changes. Authority framing exhibited the lowest TRR and highest BPI at the one-week follow-up, whereas the control condition showed the highest TRR and lowest BPI.

\end{document}